\documentclass{article}
\pdfoutput=1   % for ensuring pdf processing for arxiv
\usepackage[preprint]{nips_2018}

\usepackage[utf8]{inputenc} % allow utf-8 input
\usepackage[T1]{fontenc}    % use 8-bit T1 fonts
\usepackage{hyperref}       % hyperlinks
\usepackage{url}            % simple URL typesetting
\usepackage{booktabs}       % professional-quality tables
\usepackage{amsfonts}       % blackboard math symbols
\usepackage{nicefrac}       % compact symbols for 1/2, etc.
\usepackage{microtype}      % microtypography
\usepackage{diagbox} %backslashbox uses it for a diagonally split table cell
\usepackage{dsfont}
\usepackage{bm} % math bold. Example of use: %$\bm{BWT^{+}}$
\usepackage{color, colortbl} %cell table colouring
\usepackage{multicol}
\usepackage{amssymb}
\definecolor{Gray}{gray}{0.9}
\definecolor{LightCyan}{rgb}{0.88,1,1}  % allows to colour a row \rowcolor{LightCyan}

\usepackage[usenames,dvipsnames]{xcolor}
\usepackage{tkz-kiviat,numprint,fullpage} 
\usepackage{pgfplotstable} 
\usetikzlibrary{arrows}
\usepackage[colorinlistoftodos]{todonotes}
\title{Don't forget, there is more than forgetting: new metrics for Continual Learning}

\author{
Natalia Díaz-Rodríguez\thanks{Both authors contributed equally to this work.% \url{www.continualAI.org}
} \\
ENSTA ParisTech + INRIA Flowers, France. \\
\texttt{natalia.diaz@ensta-paristech.fr}\\
\And Vincenzo Lomonaco$^*$\\ %$^1$ \\
University of Bologna, Italy.\\
\texttt{vincenzo.lomonaco@unibo.it}
\AND David Filliat \\ %$^2$, David Filliat$^1$, Davide Maltoni$^2$ %\thanks{www.continualAI.org } Use footnote for providing further information about author (webpage, alternative    address)---\emph{not} for acknowledging funding agencies. %$^1$ U2IS, 
ENSTA ParisTech + INRIA Flowers, France. \\
\texttt{david.filliat@ensta-paristech.fr}\\
\And Davide Maltoni \\
University of Bologna, Italy.\\ 
\texttt{davide.maltoni@unibo.it}
}

\begin{document}

\maketitle

\begin{abstract}
% v0 Continual learning consists of algorithms that learn about the external world continuously and adaptively thought time, and enable the autonomous incremental development of ever more complex skills and knowledge.
% The lack of consensus in evaluating strategies, metrics, memory and operational resource requirements by continual learning algorithms motivates us to introduce more comprehensive metrics capable of ranking continual learning algorithms accounting for a desirable set of properties in continual learning. By raising concern about the importance of not only assessing how models forget (which is paramount to continual learning), we also assess other mid-term dependencies through time, expand on backward and forward transfer of knowledge, knowledge sharing, and memory and operational resource constraints. These have, we believe, practical implications worth considering in the deployment of real AI systems that learn continually.
% \cvin{I propose to modify the abstract as below:} V1: better:
Continual learning consists of algorithms that learn from a stream of data/tasks continuously and adaptively thought time, enabling the incremental development of ever more complex knowledge and skills.
The lack of consensus in evaluating continual learning algorithms and the almost exclusive focus on forgetting motivate us to propose a more comprehensive set of implementation independent metrics accounting for several factors we believe have practical implications worth considering in the deployment of real AI systems that learn continually: accuracy or performance over time, backward and forward knowledge transfer, memory overhead as well as computational efficiency.
 %propose to apply the 
Drawing inspiration from the standard Multi-Attribute Value Theory (MAVT) we further propose to fuse these metrics into a single score for ranking purposes and we evaluate our proposal with five continual learning strategies on the iCIFAR-100 continual learning benchmark.

\end{abstract}

%%%%%%%%%%%%%%%%%%%%%%%%%%%%%%%%%%%%%%%%%%%%%%%%%%%%%%%%%%%%%%%%%%%%%%%%%%%%%%%%%%%%%%%%%%%%%%%%%%%%
%%%%%%%%%%%%%%%%%%%%%%%%%%%%%%%%%%%%%%%%%%%%%%%%%%%%%%%%%%%%%%%%%%%%%%%%%%%%%%%%%%%%%%%%%%%%%%%%%%%%

%								SECTION : Introduction

%%%%%%%%%%%%%%%%%%%%%%%%%%%%%%%%%%%%%%%%%%%%%%%%%%%%%%%%%%%%%%%%%%%%%%%%%%%%%%%%%%%%%%%%%%%%%%%%%%%%
%%%%%%%%%%%%%%%%%%%%%%%%%%%%%%%%%%%%%%%%%%%%%%%%%%%%%%%%%%%%%%%%%%%%%%%%%%%%%%%%%%%%%%%%%%%%%%%%%%%%

\section{Introduction and Related Work} %: Continual Learning}
%\cvin{I would change the section tile in "Introduction and Related Works"}
 %A Continual Learning algorithms is expected to learn from a continuous stream of data only available over time. %On the contrary, non-continual learning settings assume to have access to all data at once and be able to process them as desired.
%In order to handle such non-stationarity of data streams, it is important to learn deep representations in an online manner \citep{Li16}. As data gets discarded and has a limited lifetime, the ability to forget what is not important and retain what matters for the future are the main issues that Continual Learning (CL) targets to focus on.
%\citep{Chen18} presents an overview on classic lifelong learning approaches in machine learning; the focus is more on a historical view of CL models rather than on the recent proposals on deep learning.
%CL Definition
%\subsubsection{Evaluating beyond catastrophic forgetting} 
Catastrophic forgetting %(or catastrophic interference) 
\citep{Mccloskey89,French99} refers to the well-known phenomenon %in a neural network of experiencing 
of a neural network experiencing a rapid overriding of previously learned knowledge when trained sequentially on new data \citep{Mccloskey89}. Since, by definition, continual learning algorithms deal with a sequence of data or tasks, eliminating catastrophic forgetting is an important objective often quantified for naturally assessing the quality of novel approaches \citep{Serra18,Lopez-Paz17,Hayes18,Farquhar18}.
However, the almost exclusive focus on forgetting, motivates us to propose a set of comprehensive metrics for evaluating different factors worth considering in the development of systems that learn continuously, such as accuracy or performance \emph{over time}, backward and forward knowledge transfer, memory overhead, as well as computational efficiency. For ranking purposes, we also propose a continual learning score ($CL_{score}$) which is flexible enough to be used with different weighting schemes depending on the specific application and needs. The aim of this paper is to stimulate the research community for a more methodical and comprehensive evaluation of continual learning algorithms %which bear quite novel issues 
and convey novel issues and constraints worth considering with respect to ``static'' machine learning setting.
% \cvin{For Davide, here we have to be careful: i) introduce right here that we want each metric to be between 0 and 1. ii) clarify that the cl score is optional and dependent to an arbitrary choice of context dependent weights iii) better if we stress the idea that this paper of new metrics should be considered as a proposal, a stimulus to the community to do better in evaluating CL: we should not say this is the best way of doing things.}
%\section{Related Work: Continual Learning Metrics}
%\cnat{move all related work formulas to appendix for space reasons and reduce version??}

For an exhaustive evaluation, we can assume to have access to a series of test sets $Te_i$ over time. The aim is to assess and disentangle the performance of our learning hypothesis $h_i$ as well as to evaluate if it is %and are representative for the knowledge that should be learned by the correspondent $Tr_i$.
representative of the knowledge that should be learned by the correspondent training batch $Tr_i$. However, as discussed in \citep{Lopez-Paz17}, a different granularity of the evaluation at the \textit{task} level can as well be achieved by having the same test batch for many $Tr_i$. 
% For simplicity, in the description metrics below \cite{Hayes188} assumes to have access to each $Te_i^C$, and define the \emph{cumulative training set} and \emph{cumulative test set} as $Tr_i$ and $Te_i$ respectively (see \citep{Lopez-Paz17}). 
% \begin{equation}
% Tr_{i}^{C} = \bigcup\limits_{i=1}^{i-1} Tr_{i},\ \ Te_{i}^{C} = \bigcup\limits_{i=1}^{i-1} Te_{i}.
% \end{equation}
 
 An overall performance $\Omega$ in a supervised classification setting was proposed in \citep{Hayes18} based on the relative performance of an incrementally trained algorithm with respect to an off-line trained one (which has access to all the data at once, and acts as an upper bound). %, which in our notation would be: %, with an Ωall = 1 indicating identical performance. 
% \begin{equation}
% \Omega = \frac{1}{n}\sum_{i=1}^{n}\frac{A(h_i, Te_{i}^{C})}{A(h_i^{C}, Te_{i}^{C})}. 
% \end{equation}
%Where $A$ is the accuracy measure (taking a model and a test set as input), $h_i$ is the hypothesis trained on the sequence of training batches up to the $Tr_i$ (our CL strategy) and $h_i^{C}$ is the best hypothesis we can train off-line having access to all the data in $Tr_{i}^{C}$ at once, and hence, our upper bound.
%%% LONG VERSION
% In \citep{Serra18}, instead, the authors try to directly model forgetting with the
% proposed \emph{forgetting ratio} metric $\rho$, defined as:
% \begin{equation}
% \rho = \frac{1}{n} \sum_{i=1}^{n} \left(\frac{A(h_{i-1}, Te_i^C) - \bar{b}_i}{A(h_i^C, Te_i^C) - \bar{b}_i}-1\right)
% \end{equation}
% Where, $\bar{b}$ is the vector of test accuracies for each $Te_i$ at
% random initialization (a lower bound on accuracy).
% SHORT VERSION: 
 \cite{Serra18}, instead, try to directly model forgetting with a proposed \emph{forgetting ratio} metric $\rho$ that considers a vector of test accuracies for each $Te_i$ at random initialization that represents a lower bound on accuracy. In the same setting, in \citep{Lopez-Paz17} other three important metrics are proposed: \emph{Average Accuracy} (ACC), \emph{Backward Transfer} (BWT), \emph{Forward Transfer} (FWT). In this case, after the model finishes learning about the training set $Tr_i$, its performance is evaluated on all (even future) test batches.  %:
% \begin{equation}
% ACC = A(h_n, Te_n^C)
% \end{equation}
% \begin{equation}
% BWT = \frac{1}{n-1}\sum_{i=1}^{n-1} A(h_n, Te_i) - A(h_i, Te_i)
% \end{equation}
% \begin{equation}
% FWT = \frac{1}{n-1}\sum_{i=2}^{n} A(h_n, Te_i) - \bar{b}_i.
% \end{equation}
%TODO COULD GO TO APPENDIX?
The higher these metrics, the better the model. If two models have similar ACC, the most preferable one is the one with larger BWT and FWT. %Note that it is meaningless to discuss backward transfer for the first batch, or forward transfer for the last batch.
While forgetting and knowledge transfer could be quantified and evaluated in various ways \citep{Farquhar18,Hayes18}, these may not suffice for a robust evaluation of CL strategies in practice. We thus propose a comprehensive set of metrics that expands and complements existing ones. % SPACE REASONS CUT  
%For example, in order to better understand the different properties of each strategy in different conditions, especially for embedded systems and robotics, it would be interesting to keep track and unambiguously determine the amount of computation and memory resources exploited. Stability is another important property that should be evaluated since in safety-critical conditions, potential abrupt performance drifts would be a major concern when learning continuously.   % For example, in an adversarial and generative CL setting, $P$ could be a distance based function, such as the Euclidean distance between real and generated images \citep{Seff17}. 

%%%%%%%%%%%%%%%%%%%%%%%%%%%%%%%%%%%%%%%%%%%%%%%%%%%%%%%%%%%%%%%%%%
%%%%%%%%%%%%%%%%%%%%%%%%%%%%%%%%%%%%%%%%%%%%%%%%%%%%%%%%%%%%%%%%%%
\section{Proposed Metrics for Continual Learning}
\label{sec:metric}
In Continual Learning, $\mathcal{D}$ can be thought of a potentially infinite sequence of unknown distributions $\mathcal{D} = \{D_1, \dots, D_n\}$ over $X \times Y$ we encounter over time, with $X$ and $Y$ as input and output random variables respectively. Given $h^*$ as the general target function (i.e. our ideal prediction model), a task $T$ is defined by a unique task label $t$ %$\hat{t}$ 
and its target function $g_{\hat{t}}^*(x) \equiv h^*(x,t=\hat{t})$, i.e., the objective of its learning.  % TODO: avoid speaking about distributions. Just a list of training sets. Use A-GEM h definition? Remove h and t? t is that? keep it general for other than classification tasks.
%\cdavid{Maybe introducing $h$ and $t$ in relation with $\mathcal{D}$ before the learning algorithm would be more clear.}% Done, shifted that phrase above.
A CL algorithm $A^{CL}$ is an algorithm with the following signature: 
\begin{equation}
	\forall D_i \in \mathcal{D}, \hspace{20pt} A^{CL}_i:\ \ <h_{i-1}, Tr_i, M_{i-1}, t>  \rightarrow <h_i, M_i> 
\end{equation}
Where $h_i$ is the model, $Tr_i$ is the training set of examples drawn from the respective $D_i$ distribution, $M_i$ is an external memory where we can store previous training examples and $t$ is a task label. For simplicity, we can assume $N$ as the number of tasks, one for each $Tr_i$.

In %the CL framework described above
this CL framework, %and in \citep{DiazLesortLomonaco18}, 
we propose algorithm ranking metrics according to different desiderata attainable in CL divided on seven criteria. %We use Multi-Attribute Value Theory (MAVT) \citep{Keeney93} because it is %the best 
% one of the most common ways to find a monotone function $f:[0, \infty[ \rightarrow [0,1] $,
% \cdavid{I don't undertand what you mean here by 'it is the best way to find a monotone function $f:[0, \infty[ \rightarrow [0,1] $'}\cnat{it was adviced by a math friend, it is common in operational research methods. Basically says to normalise values to a desired range by substracting the min and dividing by (max-min). I changed it to 'one of the most common ways'. Better explained now?}
In order to provide bounds to each metric (originally lying in $f:[0, \infty[$), we map it to a $[0, 1]$ range (as it is commonly done, e.g., in Multi-Attribute Value Theory (MAVT) \citep{Keeney93}) and formulate it so that its optimal value is given by its maximization. This is to preserve interpretability of the proposed aggregating $CL_{score}$ metric, and allow to evaluate CL algorithms with respect to multiple criteria, %(accuracy, memory, etc.), 
rank them from best to worst, and accommodate %criteria 
weighting schemes according to constraints and desiderata.
%These seven criteria are formulated below.

%\begin{criterion} %TODO can save space not leaving empty rest of line
\textbf{Accuracy}: 
%\end{criterion}
Given the train-test accuracy matrix $R \in \mathds{R}^{N \times N}$, which contains in each entry $R_{i,j}$ the test classification accuracy of the model on task $t_j$ after observing the last sample from task $t_i$ \citep{Lopez-Paz17}, Accuracy (A) considers the average accuracy for training set $Tr_i$ and test set $Te_j$ by considering the diagonal elements of  $R$, as well as all elements below it (see Table \ref{tab:acc-matrix-r}):

\begin{equation}
A = \frac{\sum_{i \ge j}^{N} R_{i,j}}{\frac{N(N+1)}{2}}
\end{equation}
%where $R_{i,j}$ is the accuracy matrix entry $(i, j)$ representing the accuracy of training all training sets up to $Tr_i$ (\textit{including} $Tr_i$), and testing with test set $Te_j$. 
While the A criteria was originally defined to asses the performance of the model at the end of the last task \citep{Lopez-Paz17}, we believe that an accuracy metric that takes into account the performance of the model at \emph{every timestep i in time} better characterizes the dynamic aspects of CL. The same idea is applied to the modified BWT and FWT metrics introduced below. %\footnote{Here, for simplicity we make the number of distributions \textit{n} equal to the number of tasks \textit{N}.}. %Likewise, the standard deviation of the accuracy should be reported. % todo: add formula?
%\cnat{In LopezPaz and there T is nr of tasks, however our framework definition above says n is n distributions. shall we say we make it here by simplicity we make n of distributions n equal to the nr of tasks N?}
%\begin{criterion}

%TODO long version \cdavid{Putting table 2 here seem important as you refer several times to it and it helps a lot understanding the ideas}

%\cvin{Explaining the difference from Ranzato ACC, BWT and FTW should be more delicate. In particular we can say that while this three criterion has been though as ... in this paper we think that in some contextes it may be useful to consider them at every step in time not just the end.}%DONE IN FOOTNOTE

%\begin{criterion}
\textbf{Backward Transfer} (BWT) % (Forgetting)}: % ($BWT^{-}$ and $BWT^{+}$)}: 
%\end{criterion}
%Backward transfer (BWT)
 measures the influence that learning a task has on the performance on previous tasks \citep{Lopez-Paz17}. The motivation arises when an agent needs to learn in a multi-task or data stream setting. The lifelong abilities to both improve and not degrade performance are important and should be evaluated throughout its lifetime. %to clarify, from LopezPaz: 'On the one hand, there exists positive backward transfer when learning about some task t increases the performance on some preceding task k. On the other hand, there exists negative backward transfer when learning about some task t decreases the performance on some preceding task k. Large negative backward transfer is also known as (catastrophic) forgetting'. 
It is defined as the accuracy computed on $Te_i$ right after learning $Tr_i$ as well as at the end of the last task on the same test set. (see %elements accounted from the accuracy matrix R in 
 Table \ref{tab:acc-matrix-r}% in light cyan
 ). Here, as in the accuracy metric, we expand it to consider the average of the backward transfer \emph{after each task}:
 %\footnote{Note that here we extend \citep{Lopez-Paz17} original definition of BWT and FWT to also include the intermediate sets accuracy values for a lifelong forgetting assessment, and not only the last (task) rows of train/test task set, respectively. } %\vin{a bit more of motivation here is needed}\nat{better? feel free to improve}.  %TODO: check: isnt this only for A metric and not for bwt?   \ref{tab:bwd}:
% \begin{table}[!htbp]
% \centering
% \caption{Elements accounted in the average to compute BWT criterion (\textbf{$Tr_i$} = training, \textbf{$Te_i$} = test batch)}
% \label{tab:bwd}
% \begin{tabular}{l|c|c|c} %{|p{30mm}|P{15mm}|P{15mm}|P{15mm}|}
% \hline
% % \backslashbox{\textbf{Train}}{\textbf{Test}} & \textbf{$Te_1$} & \textbf{$Te_2$} & \textbf{$Te_3$} \\\hline\hline
% R & \textbf{$Te_1$} & \textbf{$Te_2$} & \textbf{$Te_3$} \\\hline\hline
% \cellcolor{LightCyan}$Tr_1$ &  &  &   \\\hline  
% \cellcolor{LightCyan}$Tr_2$ & \cellcolor{LightCyan}$R_{ij}$ &  &   \\\hline 
% \cellcolor{LightCyan}$Tr_3$ & \cellcolor{LightCyan}$R_{ij}$ & \cellcolor{LightCyan}$R_{ij}$ &   \\\hline 
% \end{tabular}
% \end{table}
%\cnat{check with Ranzato train and test are placed in same position row/columns}
\begin{equation}
BWT = \frac{\sum_{i = 2}^{N}\sum_{j = 1}^{i-1}(R_{i,j} - R_{j,j})}{\frac{N(N-1)}{2}}
\end{equation}
% for longer version
%Applying $\frac{BWT}{100}$ transforms the original range of BWT in \citep{Lopez-Paz17} from [-100, 100] to [-1,1].
Because the original meaning of BWT assumed positive values for backward transfer and negative values to define (catastrophic) \textit{forgetting}, in order %to correct for potentially negative values in the numerator of BWT, and normalize BWT to lie in 
to map BWT to also lie on $[0, 1]$ and give more importance to two semantically different concepts, BWT is broken into two different clipped terms: % that need to be minimized and maximized, respectively
%LONG VERSION:  Footnote moved now to appendix % \footnote{The original former terms would return a domain for $BWT^{-} \in [0, 0.5)$, and for $BWT^{+} \in [0.5, 1]$, respectively which, through the clipping, are transformed, as the rest of criteria in the CL metric, to stay in [0,1].% N:this sounds too simple to say:
% %Note all accuracy metrics A, BWT, FWT are normalized to transform its original percentage range from [0, 100] to [0,1].
% }:
The originally negative (forgetting) BWT (now positive), %not forgetting, 
i.e., \textbf{Remembering}, as %, to maximize:
$REM = 1- |min (BWT, 0)|$.
%$BWT^{-} =  |min (BWT, 0)|$.
% \begin{equation}
% BWT^{-} =  1- |min (BWT, 0)|  
% \end{equation}
and (the originally positive) BWT, %, and upgraded) %\textbf{Positive 
i.e., improvement over time \textbf{Positive Backward Transfer} %, to maximize:
$BWT^{+} =  max (BWT, 0)$.
% \begin{equation}
% BWT^{+} =  max (BWT, 0)
% \end{equation}

%\begin{criterion}
\textbf{Forward Transfer}: 
%\end{criterion}
%Forward transfer (FWT) 
FWT measures the influence that learning a task has on the performance of future tasks \citep{Lopez-Paz17}. Following the spirit of the previous metrics we modify it as the average accuracy for the train-test accuracy entries $R_{i,j}$ above the principal diagonal of R, excluding it (see elements accounted in Table \ref{tab:acc-matrix-r} in light gray and additional information in the appendix). Forward transfer can occur when the model is able to perform \textit{zero-shot} learning. %\ref{tab:fwd}.
% \begin{table}[!htbp]
% \centering
% \caption{Elements accounted to compute FWT criterion (\textbf{$Tr_i$} = training, \textbf{$Te_i$}= test batch)}
% \label{tab:fwd}
% \begin{tabular}{l|c|c|c} %{|p{30mm}|P{15mm}|P{15mm}|P{15mm}|}
% \hline
% % \backslashbox{\textbf{Train}}{\textbf{Test}} & \textbf{$Te_1$} & \textbf{$Te_2$} & \textbf{$Te_3$} \\\hline\hline
% R & \textbf{$Te_1$} & \textbf{$Te_2$} & \textbf{$Te_3$} \\\hline\hline
% $Tr_1$ &  & \cellcolor{Gray}$R_{ij}$ &\cellcolor{Gray} $R_{ij}$  \\\hline  
% $Tr_2$ &  &  &  \cellcolor{Gray} $R_{ij}$ \\\hline 
% $Tr_3$ &  &  &   \\\hline 
% \end{tabular}
% \end{table}
We therefore redefine FWT as: %, i.e., the inability to forget, \nat{or ability to increase performance of a current task with future task(s),} as: %, to be maximized, as:
\begin{equation}
FWT = \frac{\sum_{i < j}^{N} R_{i,j}}{\frac{N(N-1)}{2}}
\end{equation}

%\cdavid{Why not substracing $R_{j,j}$ as in backward transfer ? Why not using positive/negative as in backward ?} \cnat{basically to avoid having to split in fwd + and - and because random initialization is not guaranteed to always be better/worse than current CL setting (see formula 4 in https://arxiv.org/pdf/1706.08840.pdf) and footnote here. True that perhaps we should split on pos and negative but since they substract something independent of matrix R, we decided to remove the substraction term (the vector of test accuracies for each task at random initialization)...there are the two options in fact, but we see no point in splitting into fwd pos and neg.}

%moved to appendix:
% The substraction term (vector $b_i$ of test accuracies for each task at random initialization) in original FWT in \citep{Lopez-Paz17} was removed to guarantee non negative values and allow for potential positive transfer as they demonstrate it is possible to happen. % as demonstrated in \citep{Lopez-Paz17}. 
% The idea is supporting the fact that algorithms can do worse than random accuracy for some strategies.
%TODO: add time criterion  Hebert Martin rate of response production. 

\textbf{Model size efficiency}: % hypothesis}
%\end{criterion}
% The memory in bits occupied by the model $\theta$, %\cnat{(use theta as in framework or h as in the code as param?)}
% $Mem(\theta)$ should be bounded in each training set, by an upper bound %equal to 
% \nat{related?} to the size of the whole lifetime dataset $D$. %\nat{However, since it is possible to have models larger than datasets (e.g. VGG on CIFAR10), the model size-dataset size ratio may not always be relevant. Model size (MS) is thus:}
The memory size of model $h_i$ quantified in terms of parameters $\theta$ at each task $i$, $Mem(\theta_i)$, should not grow too rapidly with respect to the size of the model that learned the first task, $Mem(\theta_1)$. % should be bounded with respect to the size of the model able to learn all tasks separately. 
Model size (MS) is thus:
% \begin{equation}
% MS = 1-\frac{\mathit{Mem}(\theta)}{\mathit{Mem}(D)}
% \end{equation}

% \cdavid{You should introduce $\theta$ in the begining. Comparing model size with dataset size may not always be relevant, as you can have models larger than datasets (e.g. vgg on CIFAR10). I guess Davide suggestion is related to that ? }\nat{done}
%\cvin{Davide suggests this variation in the formula:}
% \begin{equation}
% MS = 1-min(1, \mathit{avg}(\frac{Mem(\theta_i) - Mem(\theta_0)}{Mem(\theta_0) \cdot N}))
% \end{equation}
%Improved version to be more accurate: sum instead of average
% \begin{equation}
% MS = 1-min(1, \mathit{avg}(\frac{Mem(\theta_i) - Mem(\theta_0)}{Mem(\theta_0) \cdot N}))
% \end{equation}
\begin{equation}
MS = min(1, \frac{\sum_{i = 1}^{N}\frac{Mem(\theta_1)}{Mem(\theta_i)}}{N})
\end{equation}
%where $Mem(x)$ gives the memory occupation of x and N is the number of tasks. % by the parameters $\theta$ of the CL algorithm and the dataset D, respectively.

%\begin{criterion}
\textbf{Samples storage size efficiency}:  %Memory size}
%\end{criterion}
Many CL approaches save training samples % or generative replay generated samples 
as a replay strategy to not forget. The memory occupation in bits by the samples storage memory $M$, $\mathit{Mem}(M)$, %in the CL algorithm, 
should be bounded by the memory occupation of the total number of examples encountered at the end of the last task, i.e. the cumulative sum of $Tr_i$ here defined as the lifetime dataset $D$ (associated to the set of all distributions $\mathcal{D}$). Thus, we define Samples Storage Size (SSS) efficiency as: % $[0, \inf] \rightarrow [0, 1]$

% \begin{equation}
% SSS = 1-\frac{\mathit{Mem}(M)}{\mathit{Mem}(D)}
% \end{equation}
%\cvin{Davide suggests this variation in the formula:}
% \begin{equation}
% SSS = 1-min(1, \mathit{avg}(\frac{\mathit{Mem}(M_i)}{\mathit{Mem}(D)}))
% \end{equation}
\begin{equation}
%using more accurate aggregation, ie. sum in numerator:
SSS = 1-min(1, \frac{\sum_{i = 1}^{N} \frac{\mathit{Mem}(M_i)}{\mathit{Mem}(D)}}{N})
% using avg in numerator:
%SSS = 1-min(1, \frac{\frac{\sum_{i = 1}^{N}(\mathit{Mem}(M_i)}{N}}{\frac{N(N-1)}{2}} %\frac{\mathit{Mem}(M_i)}{\mathit{Mem}(D)}))
\end{equation}

%\begin{criterion}
\textbf{Computational efficiency}: 
%\end{criterion}
Since the computational efficiency (CE) is bounded by the number of multiplication and addition operations for the training set $Tr_i$, % long version i.e., $\mathit{Ops}(A^{CL}) \approx \mathit{Mul-Adds}(Tr_i)$, 
we can define the average CE across tasks as: %training sets in terms of the number of operations needed to process a training set $\mathit{Ops}(Tr_i)$ and the whole dataset $D$, $\mathit{Ops}(D)$, as:  % do we need to refer to Te here too? TODO for the production rate extra metric, relate Tri and Tei

% \begin{equation}  % TODO Nat: transform Avg to sum over N (n of training sets) to look better?
% CE = 1-\frac{\mathit{Avg}(\mathit{Ops}(Tr_i))}{\mathit{Ops}(D)}
% \end{equation}
% TODO: substracting naïve or putting in denominator makes it non portable, depends on implementation. If we put one fwd and bwd pass,  ('from scratch'), nr iterations need to be reported-> concurrency of publications problem, and portability problem.
%\cvin{Davide suggests this variation in the formula:}
% \begin{equation}
% CE = 1-min(1, \mathit{Avg}(\frac{Ops(Tr_i)}{Ops_{fs}(Tr_i)}))
% \end{equation}
% Improved version not requiring loose 'convergence' definition but requiring reporting epochs
% \begin{equation}
%% % CE = 1-min(1, \mathit{Avg}(\frac{Ops(Tr_i)}{Ops \uparrow\downarrow(Tr_i)\varepsilon}))
% % without epsilon
% CE = min(1, \frac{\sum_{i = 1}^{N} \frac{\mathit{Ops}\uparrow\downarrow(Tr_i)}{\mathit{Ops}(Tr_i)}}{N})
% \end{equation}
\begin{equation} % with epsilon:
CE = min(1, \frac{\sum_{i = 1}^{N} \frac{\mathit{Ops}\uparrow\downarrow(Tr_i) \cdot \varepsilon}{\mathit{Ops}(Tr_i)}}{N})
\end{equation}

where $Ops(Tr_i)$  is the number of (mul-adds) operations needed to learn $Tr_i$, and $Ops\uparrow\downarrow$($Tr_i$) is the number of operations required to do one forward and one backward (backprop) pass on $Tr_i$. %\vin{where $Ops_{fs}$ is the number of mul-adds needed to learn the training set $Tr_i$ from scratch}\nat{\footnote{until convergence for a reported iterations/epochs fixed reported parameter}}.
When the value of $Ops\uparrow\downarrow$($Tr_i$) is negligible w.r.t. $Ops(Tr_i)$, a scaling factor associated to the number of epochs needed to learn $Tr_i$, $\varepsilon$ %(i.e., $Ops\uparrow\downarrow$($Tr_i) \cdot \varepsilon$ in the numerator) 
larger than a default value of 1, can be used to make CE more meaningful (i.e. avoiding compression of the values very near to zero). Since we are essentially moving the lower bound of the computation, which depends on the benchmark complexity, this adjustment also translates on better interpretability of CE (Fig. \ref{fig:spider-and-plot})%\footnote{In our classification experiments we found $\varepsilon$ = 10 to improve the visual interpretability of CE (Fig. \ref{fig:spider-and-plot}.}. %; otherwise, $\varepsilon$ can default to be 1.

%it's not only about visualization but to make the CE metric more meaningful since we are essentially moving the lower bound of computation which depends on the benchmark complexity

%%%%%%%%%%%%%%%%%%%%%%%%%%%%%%%%%%%%%%%%%%%%%%%%%%%%%%%%%%%%%%%%%%%%%%%%%%%%%%%%%%%%%%%%%%%%%%%%%%%%
%%%%%%%%%%%%%%%%%%%%%%%%%%%%%%%%%%%%%%%%%%%%%%%%%%%%%%%%%%%%%%%%%%%%%%%%%%%%%%%%%%%%%%%%%%%%%%%%%%%%

%TODO long version: say it here (now in appendix) \cvin{Here again Davide suggest to point out that the CL score is optional and context dependent like the mAP metric can be useful but on a community agreement or in specific setting where the weights are clearly definable.}\cnat{ added clarification in appendix due to space}

In order to assess a CL algorithm $A^{CL}$, % with all CL metrics described, 
following \citep{Ishizaka13}, each criterion $c_i \in \mathcal{C}$ (where $c_i \in [0, 1]$) is assigned a weight $w_i \in [0,1]$ where $\sum_i^{\mathcal{C}}w_i=1$. Each $c_i$ should be the average of $r$ runs. Therefore, the final \textbf{$CL_{score}$} to maximize is computed as:
\begin{equation}
CL_{score} = %\frac{
\sum_{i=1}^{\#\mathcal{C}}w_i c_{i}   %}{\#\mathcal{C}}
\end{equation}
where each criterion $c_i$ that needs to be minimized %("-" in first column in Table \ref{tab:criteria}), 
is transformed to $c_i = 1 - c_i$ to preserve increasing monotonicity of the metric (for overall maximization of all criteria in $\mathcal{C}$). % and keep each criterion interpretable). The $CL_{stability}$ is the average of the standard deviations from all previous criteria $c_i$:
\textbf{$CL_{stability}$} is thus:
\begin{equation}
CL_{stability} = 1- \sum_{i=1}^{\#\mathcal{C}} w_i \mathit{stddev}(c_{i})  % TODO: write sigma_c_i when more space to say where sigma stands for
\end{equation}

% \cvin{I don't think if we do not have example it is worth saying this below..}\nat{I agree its quite useless, lets remove in this short version}%TODO add if we use it only
% When more than one benchmark/dataset is reported, the CL metric can average these metrics for each $b$ $\in \mathcal{B}$ where $\mathcal{B}$ is the set of evaluation benchmarks assessed by $A^{CL}$. 

%		SECTION : Experiments and results
%%%%%%%%%%%%%%%%%%%%%%%%%%%%%%%%%%%%%%%%%%%%%%%%%%%%%%%%%%%%%%%%%%%%%%%%%%%%%%%%%%%%%%%%%%%%%%%%%%%%
%%%%%%%%%%%%%%%%%%%%%%%%%%%%%%%%%%%%%%%%%%%%%%%%%%%%%%%%%%%%%%%%%%%%%%%%%%%%%%%%%%%%%%%%%%%%%%%%%%%%

\section{Experiments and Conclusions}

\begin{figure}[htbp!]
\centering
\includegraphics[height=6cm]{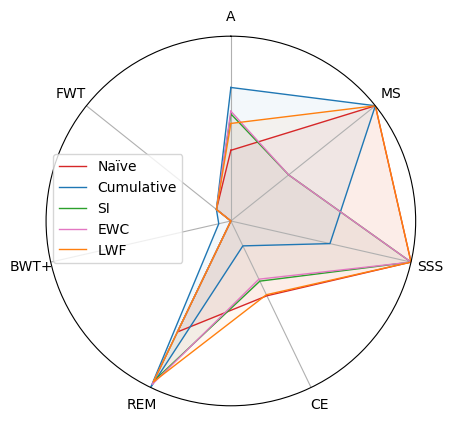}
\includegraphics[height=5.8cm]{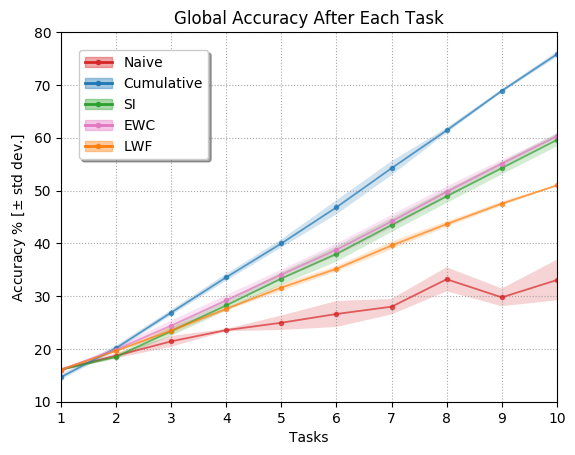}%\\ \hline
 % \end{tabular} 
  \caption{a) Spider chart: CL metrics per strategy (larger area is better). %\cnat{here put all metrics together and in appendix add one per one with min and max bound}.
  b) Accuracy per CL strategy computed over the fixed test set as proposed in \citep{Lomonaco17, Maltoni18}. %TODO long vers [$\pm$ std dev].
  }\label{fig:spider-and-plot}
\end{figure}

%TODO IN LONG VERSION: \cvin{For the table Davide suggest to split the criterion from the CL score (two separate table), to better mark the fact that only for our weights arbitrary setting EWC is doing better.}\cnat{I agree, but for this version it would mean adding a double extra column for each strategy (if in the same row/space to be on the main paper and not appendix so I propose doing it for next longer version, also it is easy to see final aggregated metrics in comparison with the partial ones this way all in one line. I added a line to make clear they are of different category!}

%To graphically and quantitatively illustrate the new metrics, we test the proposed CL metric scores on the latest and most popular continual learning strategies as well as on two application scenarios, on an object recognition task, as well as an 2D RL navigation task. 
%\subsection{Benchmarks: iCIFAR-100, CORe50 Dataset and VizDoom environment}
%\textbf{Benchmark}: %iCIFAR-100}:  
%\cnat{I'd say would be nice min 3 CL benchmarks to evaluate on, the more diverse among them, the better}
We evaluate the CL metrics on cumulative and naïve baseline strategies as in \citep{Maltoni18}, % (upper-bound),  %(lower), % and three CL strategies: 
Elastic Weight Consolidation (EWC) \citep{Kirkpatrick16}, Synaptic Intelligence (SI) \citep{Zenke17} and Learning without Forgetting (LwF) \citep{Li16}%\footnote{A notebook implementation of the metrics is available at \url{https://github.com/ContinualAI/colab/blob/master/notebooks/cl_metrics.ipynb}}
. %The latter three correspond to three different strategies from the regularization family of approaches.%Todo cite maltoni in long version?   %GEM \citep{Lopez-Paz17}, iCaRL \citep{Rebuffi18} and AR1 \citep{Maltoni18}. 
%The datasets we use include a continual object recognition task with CORe50 dataset \citep{Lomonaco17} in a single incremental task (SIT) setting. We also evaluate on VizDoom 2D environment \citep{Kempka16} in an RL task where the agent, in 3 different rooms, needs to pick fruits and avoid poisonous objects. 
iCIFAR-100 \citep{Rebuffi18} dataset %(100 classes) 
is used: each task consists of a training batch of 10 (disjoint) classes at a time. %In [20], the authors decided to further split the iCIFAR-100 benchmark in 20 different “tasks” containing 5 different classes each.
%Fig \ref{fig:spider-and-plot}a shows in a spider chart all criteria re-scaled for visibility. \ref{fig:spider-and-plot}b

Results for each proposed metric %on the test set with standard deviations are
are illustrated in %Appendix 
Table \ref{tab:benchmark} (each criterion $c_i$ reports the average over 3 runs); Fig. \ref{fig:spider-and-plot} %shows the accuracy metric in a per task basis, permitting to observe the variability on each CL metric ranking. We 
illustrates the CL metrics variability for each criterion reflecting a desirable property of CL algorithms, %reflecting a CL desideratum, %. % the fact the numbers as so near is because we want the metrics to be portable across experiments
%Given the variability on each metric ranking, we demonstrate 
as well as the needs of novel techniques addressing different aspects than accuracy and forgetting, that can be important depending on the application. For simplicity, we chose an homogeneous configuration of criteria weights that values each CL metric equally (i.e., each $w_i = \frac{1}{\#\mathcal{C}}$). However, the Appendix shows results on other possible configurations. 

While the $CL_{score}$ is optional to report, the aim of the metrics and results is to stimulate comprehensive evaluation practices. In future work we plan to refine these metrics and assess more strategies in more exhaustive evaluation settings.

\begin{table}[!htbp]
\centering
% LONG VERSION \caption{Elements in R accounted to compute the Accuracy criterion are in white and light gray. $R^{*} = R_{ii}$, \textbf{$Tr_i$} = training, \textbf{$Te_i$}= test batch}
 \caption{CL metrics and $CL_{score}$ for each CL strategy evaluated (higher is better).}
\label{tab:benchmark}
\begin{tabular}%{lccccccccc} 
{p{9mm}p{8mm}p{8mm}p{8mm}p{8mm}p{8mm}p{8mm}p{8mm}|p{14.4mm}p{19mm}}
% & \textbf{$Te_1$} & \textbf{$Te_2$} & \textbf{$Te_3$} \\\hline\hline  % TODO: Just to save space, revert to split cell in long version
%IT SHOWS UGLY splitting the cell so I removed it...: \backslashbox{\textbf{Strategy}}{\textbf{Metric}}  
\textbf{Strategy} & \textbf{A} & \textbf{REM} & \textbf{BWT\textsuperscript{+}} & \textbf{FWT} & \textbf{MS} & \textbf{SSS} & \textbf{CE} &  $\bm{CL_{score}}$  & $\bm{CL_{stability}}$ \\%\textbf{$CL_{stability}$}\\
\specialrule{.1em}{.05em}{.05em} 
%TODO make superindex in text mode to make it look same font? I dont know how without using math mode
% Uses Ranzato acc
% \textbf{Naïve} & 0.3304 & \textbf{1.0000} & \textbf{1.0000} & \textbf{0.4492} & 0.6664 & 0.0000 & 0.1000 & 0.5066 & 0.9939 \\%\hline
% \textbf{Cumul.} & \textbf{0.7581} & \textbf{1.0000} & 0.5500 & 0.1496 & \textbf{1.0000} & \textbf{0.0673} & 0.1000 & 0.5178 & 0.9983 \\%\hline
% \textbf{EWC} & 0.6029 & 0.4000 & \textbf{1.0000} & 0.3495 & 0.9821 & 0.0000 & 0.1000 & 0.4906 & 0.9987 \\%\hline
% \textbf{LWF} & 0.5097 & \textbf{1.0000} & \textbf{1.0000} & 0.4429 & 0.9667 & 0.0000 & 0.1000 & \textbf{0.5742} & \textbf{0.9994} \\%\hline
% \textbf{SI} & 0.5958 & 0.4000 & \textbf{1.0000} & 0.3613 & 0.9620 & 0.0000 & 0.1000 & 0.4884 & 0.9978 \\

\textbf{Naïve} & 0.3825 & 0.6664 & 0.0000 & 0.1000 & \textbf{1.0000} & \textbf{1.0000} & \textbf{0.4492} & 0.5140 & \textbf{0.9986} \\ 
\textbf{Cumul.} & \textbf{0.7225} & \textbf{1.0000} & \textbf{0.0673} & 0.1000 & \textbf{1.0000} & 0.5500 & 0.1496 & 0.5128 & 0.9979 \\ 
\textbf{EWC} & 0.5940 & 0.9821 & 0.0000 & 0.1000 & 0.4000 & \textbf{1.0000} & 0.3495 & 0.4894 & 0.9972 \\
\textbf{LWF} & 0.5278 & 0.9667 & 0.0000 & 0.1000 & \textbf{1.0000} & \textbf{1.0000} & 0.4429 & \textbf{0.5768} & \textbf{0.9986} \\
\textbf{SI} & 0.5795 & 0.9620 & 0.0000 & 0.1000 & 0.4000 & \textbf{1.0000} & 0.3613  & 0.4861 & 0.9970 \\
\specialrule{.1em}{.05em}{.05em} 
\end{tabular}
\end{table}

\section{Acknowledgements}
We acknowledge Matteo Brunelli and Dmitry Ponomarev for their brainstorm support.% advice on Operational Research methods. %and the  DREAM project\footnote{\url{http://www.robotsthatdream.eu}} through the European Union Horizon 2020 FET research and innovation program under grant agreement No 640891.
\bibliographystyle{apalike}
\bibliography{all_refs}

\newpage

\appendix
\section{Implementation Details}

%%%%%%%%%%%%%%%%%%%%%%%%%%%
\subsection{Metrics and benchmark illustration }

\begin{table}[!htbp]
\centering
 \caption{Elements in $R$ accounted to compute the Accuracy (white and cyan elements), BWT (in cyan), and FWT (in light gray) criteria. %\nat{Normally optimal} 
 $R^{*} = R_{ii}$, \textbf{$Tr_i$} = training, \textbf{$Te_i$}= test tasks.}
\label{tab:acc-matrix-r}
\begin{tabular}{l|ccc} %{|p{30mm}|P{15mm}|P{15mm}|P{15mm}|}
%\backslashbox{\textbf{Train}}{\textbf{Test}} & \textbf{$Te_1$} & \textbf{$Te_2$} & \textbf{$Te_3$} \\\hline\hline  % TODO: Just to save space, revert to split cell in long version
\textit{R}  & \textbf{$Te_1$} & \textbf{$Te_2$} & \textbf{$Te_3$} \\
\specialrule{.1em}{.05em}{.05em} 
\textbf{$Tr_1$} & $R^{*}$ & \cellcolor{Gray}$R_{ij}$ & \cellcolor{Gray}$R_{ij}$  \\ 
\textbf{$Tr_2$} & \cellcolor{LightCyan}$R_{ij}$ &$R^{*}$ & \cellcolor{Gray}$R_{ij}$  \\
\textbf{$Tr_3$} & \cellcolor{LightCyan}$R_{ij}$ & \cellcolor{LightCyan}$R_{ij}$ & $R^{*}$ \\
\specialrule{.1em}{.05em}{.05em} 
\end{tabular}
\end{table}

%% Repeated paragraph but put  here for context to the matrix:
Matrix $R \in \mathds{R}^{N \times N}$ contains in each entry $R_{i,j}$ the test classification accuracy of the model on task $j$ %$t_j$ 
after observing the last sample from task $i$ %$t_i$ 
\citep{Lopez-Paz17}. $N$ is the number of tasks; here for simplicity we make the number of distributions $n$ equal to $N$.  Table \ref{tab:acc-matrix-r} shows the elements in the accuracy matrix used for each metric for an example matrix of $N = 3$ tasks. $R^{*} = R_{ii}$ coincides with the (normally) optimal accuracy right \textit{after} using training set $Tr_i$ and testing on test set $Te_i$. 

Note that in order to compute Accuracy, we do not only consider as \citep{Lopez-Paz17} the last row of the accuracy matrix $R$, but also steps in between each new training set learned, to acknowledge the degradation and improvement through every time step in time.

In FWT, the substraction term (vector $b_i$ of test accuracies for each task at random initialization) in the original FWT formula in \citep{Lopez-Paz17} was removed in our definition of FWT in order to guarantee non negative values (i.e. in case of negative FWT) and allow for potential positive transfer, as they demonstrate it is possible to happen with a shared output space. % as demonstrated in \citep{Lopez-Paz17}.
The idea is supporting the fact that algorithms can do worse than random accuracy for some strategies (we refer the reader to \citep{Lopez-Paz17} for cases of positive FWT).

% \nat{When averaging metrics over each distribution or task i $\in$ [1, \#$\mathcal{D}$],  either mean or max value of the metric (e.g., in SSS) can be reported, as long as it is explicit, depending on the variability and strategy use case.}\cnat{ NOTE:Is this leaving too much freedom, should we choose max, average with std error or leave it a bit free?}. In this case, an \nat{average} %the metrics MS and SSS, and an avg to average the CE.
% was used to aggregate metrics over each task $i$.

The original BWT \citep{Lopez-Paz17} %former terms 
would return domains for $BWT^{-} \in [0, 0.5)$, and for $BWT^{+} \in [0.5, 1]$, respectively which, through the clipping, are transformed, as the rest of criteria in the CL metric, to stay in [0,1].% N:this sounds too simple to say:

\textbf{Criteria weights setting and experiments }
Despite the experiments showing the $CL_{score}$ to be optional and context dependent; the aggregation score is most meaningful when a community agrees on a particular evaluation criteria (similarly to the mAP metric), or in specific settings where the weights for the different criteria are clearly definable. Our experiments use three weight configurations $W$ = [$w_A, w_{MS}, w_{SSS}, w_{CE}, w_{BWT}, w_{Rem}, w_{FWT}$]. The first one used homogeneous weights (each $w_i = \frac{1}{\#\mathcal{C}}$) and the second and third use $W_2$ = [0.4, 0.1, 0.1, 0.1, 0.2, 0.05, 0.05] and $W_3$ = [0.4, 0.05, 0.2, 0.2, 0.05, 0.05, 0.05], as particular examples aiming at reflecting what the recent CL literature has roughly been valuing the most; however, any configuration could be used.

\textbf{Model}:
The CNN model used in this experiment is the same used in \citep{Zenke17} and \citep{Maltoni18} and consists of 4 convolutional + 2 fully connected layers (details available in Appendix A of \citep{Zenke17}).   %\cvin{the model is not pre-trained for these exps.%The model was pre-trained on CIFAR-10 \citep{Krizhevsky09}. \cnat{optimized doing? ->Human Gradient Descend ;P}}
Hyper-parameters are chosen to maximize the accuracy metric $A$ for each strategy. 

\textbf{Benchmark}: 
CIFAR-100 \citep{Krizhevsky09} classification dataset has 100 classes containing 600 natural images (32$\times$32) each (500 training + 100 test). %The default classification benchmark can be translated into a SIT-NC scenario (denoted as iCIFAR-100 by [27]) %%3GoogleNet, as many modern network architectures, does not include fully connected layers before the last classification layer.
The CL setting of iCIFAR-100 splits the 100 classes in groups. In this paper we consider groups of 10 classes, and therefore obtain 10 incremental batches.

\textbf{Baselines}: 
The lower and upper bound CL strategies are naïve and cumulative learning, respectively. The naïve learning strategy starts at $Tr_1$ %with random initialization 
and learns continuously the coming training sets $Tr_2, ..., Tr_N$ simply tuning the model across batches without any specific mechanism to control forgetting, except early stopping. The cumulative strategy starts from scratch every time, learning from the accumulation of $Tr_1, ... , Tr_{i-1}, Tr_i$ retrained with the patterns from the current batch and all previous batches (only in this approach we assume that all previous data can be stored and reused). This cumulative method is a sort of upper bound, or ideal performance that CL strategies should try to reach \citep{Maltoni18}.

\begin{figure}[htbp!]
\centering
\includegraphics[width=4.5cm]{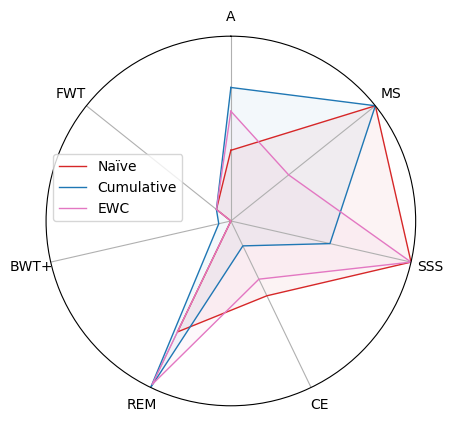} % TODO Rename for each non trivial strategy
\includegraphics[width=4.5cm]{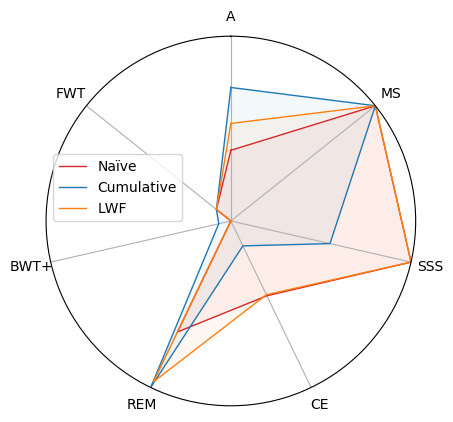}
\includegraphics[width=4.5cm]{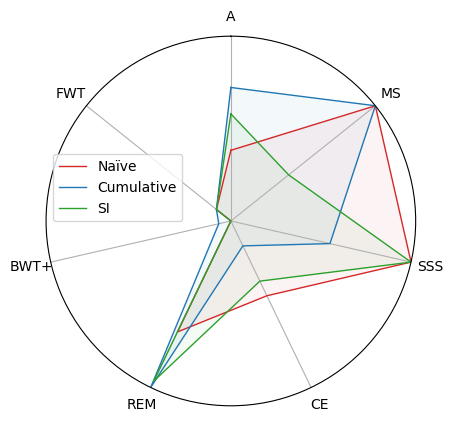}
%\\ \hline
 % \end{tabular} 
  \caption{Spider chart with CL metrics showing %(top-down, left to right) 
  CL strategies EWC, LWF and SI with their respective lower and upper bound (Naïve and Cumulative resp.) as reference baselines (to properly visualize Fig. \ref{fig:spider-and-plot}). The weight configuration for each criterion used is $W_1$ where $w_i =\frac{1}{7}$ for each $w_i \in W$.
  \label{fig:all-spider-plots}}  % W0 where wi = 1/7
\end{figure}

Spider chart in Fig \ref{fig:spider-and-plot} shows %right name? 
all objective criteria, %converted so that they are to be maximized and in this way, kept interpretable,
where the larger the area occupied under the CL algorithm curve, the highest $CL_{score}$ (more optimal) it is. % 1-ci still seems the most fair solution to make sense of it: 'the more area, the better the CL algorithm'}
Fig. \ref{fig:all-spider-plots} shows each of the main CL strategies put in context compared with the considered lower and upper bounds respectively, i.e., naïve, and cumulative strategies. The farther away the evaluated strategy is from the cumulative (blue) surface, the larger room for improvement for the CL strategy. % can aspire to.

% \begin{figure}[htbp!]
% \centering
% \includegraphics[width=4.5cm]{all_stratsW0.png} % TODO Rename for each non trivial strategy
% \includegraphics[width=4.5cm]{all_stratsW1.png}
% \includegraphics[width=4.5cm]{all_stratsW2.png}
% %\\ \hline
%  % \end{tabular} 
%   \caption{Spider chart with CL metrics showing (left to right) CL strategies Naïve, Cumulative, EWC, SI and LWF, EWC and SI with respect to .}\label{fig:all-spider-plots}
% \end{figure}

% \begin{figure}[htbp!]
% \centering
%   \begin{tabular}{ccc} 
%  %\textbf{W} % Weights Config.  & 
%  \textbf{EWC} & \textbf{SI} & \textbf{LwF} \\\hline %\\[0.0005pt]
% %$w_i = 1/7 \forall i \in c_i \in \mathcal{C}$ 
% %$W_1$: $w_i = \frac{1}{7}$ &
% \includegraphics[height=0.95in]{EWC_W1.png} & \includegraphics[height=0.95in]{SI_W1.png} & \includegraphics[height=0.95in]{LWF_W1.png} & \\ %\hline
% %Tuned $W_2$ & 
% \includegraphics[height=0.95in]{EWC_W2.png} & \includegraphics[height=0.95in]{SI_W2.png} & \includegraphics[height=0.95in]{LWF_W2.png} &  \\ 
% %Tuned $W_3$ & 
% \includegraphics[height=0.95in]{EWC_W3.png} & \includegraphics[height=0.95in]{SI_W3.png} & \includegraphics[height=0.95in]{LWF_W3.png} \\
%   \end{tabular} 
%   \caption{CL metrics for different strategies and weight configurations $W$. The first row shows an homogeneous criteria weight configuration $W_1$ where $w_i =\frac{1}{7}\forall w_i \in W$. The second row uses metric weights tuned according to a concrete weight configuration $W_2$ where [$w_A, w_{MS}, w_{SSS}, w_{CE}, w_{REM^{+}}, w_{BWT}, w_{FWT}$] = [0.4, 0.1, 0.1, 0.1, 0.2, 0.05, 0.05] (that could reflect current literature preferences). Third row shows another arbitrary configuration $W_3$ = \nat{Fix[0.4, 0.1, 0.1, 0.1, 0.2, 0.05, 0.05]}.  \label{fig:radars-different-weights}}
% \end{figure}

\begin{table}[!htbp]
\centering
\caption{$CL_{score}$ and $CL_{stability}$ for all CL strategies according to different weighting  configurations $W_i$ = [$w_A, w_{MS}, w_{SSS}, w_{CE}, w_{REM^{+}}, w_{BWT}, w_{FWT}$], where $W_1$ sets $w_i =\frac{1}{7}$ for each $w_i \in W$. The second setting of a concrete metric weights is $W_2$ = [0.4, 0.05, 0.2, 0.1, 0.15, 0.05, 0.05]. % (that could reflect current literature preferences).
A third arbitrary configuration is $W_3$ = [0.4, 0.05, 0.2, 0.2, 0.05, 0.05, 0.05].}
\label{tab:batch-examples}
\begin{tabular}{l|ccc|ccc} %{|p{30mm}|P{15mm}|P{15mm}|P{15mm}|}
\textbf{Strategy/CL Metric} 
&\multicolumn{3}{c}{$\bm{CL_{score}}$ }  & \multicolumn{3}{c}{$\bm{CL_{stability}}$} \\\hline
& \textbf{$W_1$} & \textbf{$W_2$} & \textbf{$W_3$} & \textbf{$W_1$} & \textbf{$W_2$} & \textbf{$W_3$}\\
\specialrule{.1em}{.05em}{.05em} 
\textbf{Naïve} & 0.5140 & 0.5529 & 0.5312 & \textbf{0.9986} & 0.9969 & \textbf{0.9973} \\ 
\textbf{Cumulative} & 0.5128 & 0.6223 & 0.5373 & 0.9979 & 0.9976 & 0.9964 \\
\textbf{EWC} & 0.4894 & 0.6449 & 0.5816 & 0.9972 & 0.9976 & 0.9940 \\
\textbf{LWF} & \textbf{0.5768} & \textbf{0.6554} & \textbf{0.6030} & \textbf{0.9986} & \textbf{0.9990} & 0.9972\\
\textbf{SI} & 0.4861 & 0.6372 & 0.5772 & 0.9970 & 0.9945 & 0.9927\\
\specialrule{.1em}{.05em}{.05em} 
\end{tabular}
\end{table}

\end{document}